\documentclass[11pt,a4paper]{article}
\usepackage[hyperref]{acl2021}
\usepackage{times}
\usepackage{latexsym}

\usepackage{microtype}

\aclfinalcopy 
\def\aclpaperid{3593} 

\usepackage{graphicx}
\usepackage{amsmath}
\usepackage{amssymb}
\usepackage{multirow}
\usepackage{array}
\usepackage{algorithm}
\usepackage{algorithmic}
\usepackage{enumitem}
\usepackage{colortbl}
\usepackage{makecell}

\title{A Training-free and Reference-free Summarization Evaluation Metric via Centrality-weighted Relevance and Self-referenced Redundancy}

\author{Wang Chen\textsuperscript{\rm 1}\thanks{\enspace This work was mainly done when Wang Chen was an intern at Tencent AI Lab.} \qquad
Piji Li\textsuperscript{\rm 2} \qquad
Irwin King\textsuperscript{\rm 1} \\
\textsuperscript{\rm 1}Department of Computer Science and Engineering,\\
The Chinese University of Hong Kong, Shatin, N.T., Hong Kong \\
\textsuperscript{\rm 2}Tencent AI Lab\\
{\tt \textsuperscript{\rm 1}\{wchen, king\}@cse.cuhk.edu.hk}\\
{\tt \textsuperscript{\rm 2}pijili@tencent.com}
}

\date{}

\begin{document}
\maketitle
\begin{abstract}
In recent years, reference-based and supervised summarization evaluation metrics have been widely explored. However, collecting human-annotated references and ratings are costly and time-consuming. To avoid these limitations, we propose a training-free and reference-free summarization evaluation metric. Our metric consists of a centrality-weighted relevance score and a self-referenced redundancy score. The relevance score is computed between the pseudo reference built from the source document and the given summary, where the pseudo reference content is weighted by the sentence centrality to provide importance guidance. Besides an $F_1$-based relevance score, we also design an $F_\beta$-based variant that pays more attention to the recall score. As for the redundancy score of the summary, we compute a self-masked similarity score with the summary itself to evaluate the redundant information in the summary. Finally, we combine the relevance and redundancy scores to produce the final evaluation score of the given summary. Extensive experiments show that our methods can significantly outperform existing methods on both multi-document and single-document summarization evaluation. The source code is released at \href{https://github.com/Chen-Wang-CUHK/Training-Free-and-Ref-Free-Summ-Evaluation}{https://github.com/Chen-Wang-CUHK/Training-Free-and-Ref-Free-Summ-Evaluation}.
\end{abstract}

\section{Introduction}
Text summarization systems have been developed rapidly due to the appearance of sequence-to-sequence frameworks~\cite{DBLP:conf/nips/SutskeverVL14_s2s,DBLP:journals/corr/BahdanauCB14,see-etal-2017-get_PGNet,DBLP:conf/sigir/ChanCK20}, transformer architectures~\cite{DBLP:conf/nips/VaswaniSPUJGKP17_transformer} and large-scale pre-training models~\cite{devlin-etal-2019-bert-tmp,DBLP:journals/corr/abs-1907-11692_roberta}. How to accurately evaluate the summaries generated from these systems also attracts more and more attention in this research area. One of the most accurate evaluation methods is human evaluation. However, human evaluation is expensive, time-consuming, and non-reproducible. Thus, it is necessary to develop automatic evaluation metrics for text summarization systems. Existing automatic summarization evaluation metrics can be roughly categorized into two groups: reference-based metrics and reference-free metrics. In this work, we focus on reference-free metrics.


Reference-free summarization evaluation metrics have been developed in parallel in multi-document summarization and single-document summarization. The SOTA reference-free method for multi-document summarization evaluation, SUPERT~\cite{DBLP:conf/acl/GaoZE20_supert}, predicts a relevance score for each (document, summary) pair to estimate the informativeness of the summary and then averages all the scores from multiple documents as the final evaluation score. For each pair, SUPERT employs the top-ranked sentences which are ranked by the position or centrality as a pseudo reference of the document and then applies BERTScore~\cite{DBLP:conf/iclr/ZhangKWWA20_bertscore} to produce a relevance score between the pseudo reference and the given summary. The SOTA single-document summarization reference-free evaluation metric, LS\_Score~\cite{DBLP:journals/corr/abs-2010-01781_ls_score}, combines a learned linguistic scorer for the summary and a cosine similarity scorer for the (document, summary) pair to produce the final score.

Although SUPERT and LS\_Score achieve the SOTA performance on their own areas respectively, they still have several drawbacks. For example, SUPERT only considers the relevance score between the document and the summary while ignoring the other aspects such as how much redundant information is contained in the summary. Besides, SUPERT assumes that all pseudo reference sentences are equally-important. However, in the real world, the key information of a document is unevenly distributed over sentences. Therefore, such an assumption may introduce extra noise for the evaluation.
Note that although SUPERT may employ sentence centrality to select document sentences as a pseudo reference, they ignore the sentence centrality after the selection and still treat the selected sentences equally-important.
As for LS\_Score, although it does not require a reference during the evaluation of a summary, it requires a large-scale training dataset with reference summaries to train the linguistic scorer. 
Besides the intrinsic drawbacks in these SOTA methods, to our best knowledge, there is no reference-free evaluation metric showing that it can achieve the SOTA performance on both multi-document and single-document summarization.

To solve the above limitations, based on SUPERT, we propose a novel training-free and reference-free metric for both multiple and single document summarization evaluation. Our metric is composed of a centrality-weighted relevance score and a self-referenced redundancy score. 

For the relevance score which is employed to estimate the informativeness of the summary, we incorporate the following new features. First, unlike previous work which only utilizes the token-level representations, motivated by \citet{DBLP:conf/acl/ClarkCS19_s_wms}, we engage a hybrid way that contains both token-level representations and sentence-level representations to encode the document and the summary. The purpose of the hybrid representation is to enable our method to consider richer mapping styles (i.e., token-to-token, sentence-to-token, and sentence-to-sentence) and help to produce a more comprehensive evaluation score. Second, we utilize the sentence centrality computed from sentence-level representations of the source document to produce the importance weights of the pseudo reference sentences and tokens. Based on the weights, we compute a weighted relevance score that is more precise by considering the relative importance. Third, besides the $F_1$ version of our relevance score, we also propose an adaptive $F_\beta$ version where recall is considered $\beta$ times as important as precision. $\beta$ is computed based on the length ratio between the pseudo reference and the given summary. The motivation is to punish the short summary that can easily get high precision while covering very limited important information in the pseudo reference (i.e., low recall).

To measure the redundancy of a summary, we design a simple but effective self-referenced similarity score. If a summary contains much redundant information, there must exist plenty of semantically similar tokens or sentences. Based on this assumption, we use the summary itself as the reference and input a (summary, summary) pair into a self-masked BERTScore to produce a redundancy score that evaluates the averaged degree of semantic similarity of each token or sentence with other tokens or sentences. 

After obtaining the centrality-weighted relevance score and the self-referenced redundancy score, we combine them to predict the final evaluation score. Depending on either $F_1$ or $F_\beta$ is applied in our relevance score, we propose two variants of our method: the $F_1$-based version and the $F_\beta$-based version. Extensive experiments are conducted on both multi-document and single-document summarization datasets. The results show that our $F_1$-based method already outperforms all the SOTA baselines on all datasets. Moreover, our $F_\beta$-based method can further improve the performance on multi-document summarization datasets.

Our contributions are summarized as follows: (1) A novel training-free and reference-free summarization evaluation metric which considers both relevance and redundancy; (2) A centrality-weighted relevance score that effectively utilizes the sentence centrality of the documents to provide importance guidance for the pseudo reference tokens and sentences. Besides the $F_1$ version, we also develop an $F_\beta$ based relevance score which pays more attention to recall; (3) A self-referenced redundancy score that utilizes a self-masked BERTScore to detect the duplicated information of the given summary; (4) To the best of our knowledge, we are the first evaluation metric that can achieve SOTA performance on both multiple and single document summarization under the reference-free setting.

\section{Preliminary}
\textbf{Notations.} We denote vectors as bold lowercase characters and matrices as bold uppercase characters. The characters that are not bold are used to denote scalars. Calligraphy uppercase characters are utilized to represent sets. 

\begin{figure*}[t]
  \centering
  \includegraphics[width=\textwidth]{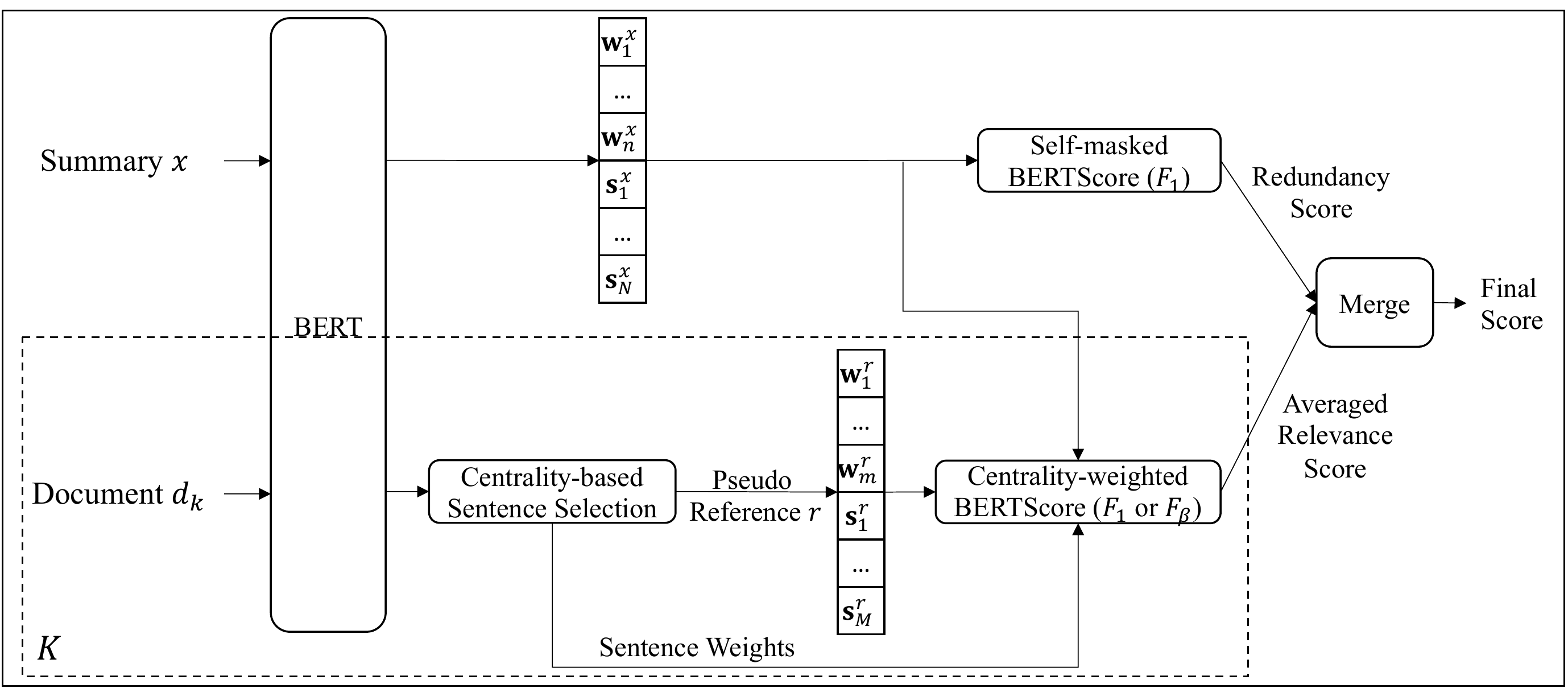}
  \caption{Overall framework of our method. $\mathbf{w}$ and $\mathbf{s}$ are the token-level and sentence-level representations. $n$ and $N$ ($m$ and $M$) are the token number and the sentence number of the summary (pseudo reference). For multi-document summary (i.e., $K > 1$), we compute  relevance scores between the summary $x$ and each document $d_k$, and then average them as the final relevance score.
  }
  \label{figure: model_framework}
\end{figure*}

\smallskip
\noindent \textbf{Problem Definition.} We formally define the reference-free summarization evaluation problem as follows. Give a set of documents $\mathcal{D}=\{d_1, d_2, ...,d_K\}$ and a generated summary $x$, the goal is to predict a score to represent the overall quality of the summary. $K=1$ and $K>1$ indicate single-document and multi-document summarization respectively.

\section{Our Methodology}
The overall framework is illustrated in Figure~\ref{figure: model_framework}. Our final evaluation score of a summary consists of an averaged centrality-weighted relevance score and a self-referenced redundancy score. Both scores are calculated on a semantic-level instead of utilizing $n$-gram overlapping. The averaged relevance score is computed from the relevance score between the summary and each document in the document set. The redundancy score is calculated based on the summary itself.
\subsection{Centrality-weighted Relevance Score}
Our relevance score aims to estimate the informativeness of the given summary. We first encode each document in the document set and the summary into hidden representations. Then, for each document, we select essential sentences by centrality to build a pseudo reference. Next, we compute a centrality-weighted relevance score between the summary and each pseudo reference. Finally, we average all the relevance scores as the final relevance score of the summary. We use the $k$-th document $d_k$ and a summary $x$ as an example to show the workflow.


\smallskip
\noindent \textbf{Encoding.} Following SUPERT~\cite{DBLP:conf/acl/GaoZE20_supert}, we first split the document $d_k$ and the summary $x$ into sentences. Then, the pre-trained SBERT\footnote{bert-large-nli-stsb-mean-tokens} is employed to encode the tokens of each sentence into token-level contextual hidden representations. We also apply max-pooling on all the tokens of a sentence to obtain the sentence-level hidden representation. Following previous work, when utilizing the token-level representations to compute the relevance and redundancy scores, we will filter out the non-informative tokens such as stop-words to improve the efficiency. 

\smallskip
\noindent \textbf{Building Pseudo Reference.} We do not choose all the document sentences of $d_k$ to evaluate the relevance of the summary. Because the whole document usually contains plenty of unimportant sentences which may introduce extra noise for the relevance evaluation. Thus, we select important document sentences to build a pseudo reference $r$ for the evaluation. The sentence selection is based on the centrality of each sentence, which is computed by the unsupervised algorithm, PacSum~\cite{DBLP:conf/acl/ZhengL19_pacsum}, using the sentence-level representation. After obtaining the centrality scores of all sentences of the document, we choose the top-$M$\footnote{In experiments, we follow the default configuration of SUPERT and set $M$ as 12 for all the datasets.} sentences as the pseudo reference. Besides, we normalize the centrality scores to $[0, 1]$ and denote the normalized centrality scores of the selected sentences as $\bar{\mathbf{a}}^s=[\bar{a}^s_1, \bar{a}^s_2, ..., \bar{a}^s_M]$ where $\bar{a}^s_i \in [0, 1]$ and the superscript $s$ means sentence-level. We denote the pseudo reference building process as \textbf{PacSumTopM}.

\smallskip
\noindent \textbf{Computing Relevance Score with One Pseudo Reference.} Instead of only using token-level representations, we also leverage the sentence-level representations to provide multi-level information. The hybrid representations of the summary $x$ and the pseudo reference $r$ are denoted as follows:
\begin{align}
    \mathbf{X} 
               &=[\mathbf{w}^x_1, ..., \mathbf{w}^x_n, \mathbf{s}^x_1, ..., \mathbf{s}^x_N], \\
    \mathbf{R}_k 
                &= [\mathbf{w}^r_1, ..., \mathbf{w}^r_m, \mathbf{s}^r_1, ..., \mathbf{s}^r_M],
\end{align}
where $n$ and $N$ ($m$ and $M$) are the token number and sentence number of the summary (pseudo reference). $\mathbf{w}$ and $\mathbf{s}$ represent the token and sentence hidden representations respectively.

Besides the hybrid representations, we also introduce a centrality weighting scheme to weight the tokens and sentences of the pseudo reference, which is different from previous work that either treats them equally or uses the surface statistics like IDF as the weights. Based on the centrality scores of the selected pseudo reference sentences i.e., $\bar{\mathbf{a}}^s=[\bar{a}^s_1, \bar{a}^s_2, ..., \bar{a}^s_M]$, we assign the weights of the pseudo reference tokens as follows:
\begin{align}
    \bar{\mathbf{a}}^w &= [\bar{a}^w_1, \bar{a}^w_2, ..., \bar{a}^w_m], \\
    \bar{a}^w_j &= \bar{a}^s_{i: w_j \in s_i},
\end{align}
where $\bar{a}_{i: w_j \in s_i}$ indicates the token $w_j$ inherits the centrality score from its sentence $s_i$. Since we have already removed the non-informative tokens in the token-level representations of each sentence, the remaining tokens capture the key information of the sentence and consequently it is reasonable to perform such a weight inheritance. Next, we combine token weights $\bar{\mathbf{a}}^w$ and sentence weights $\bar{\mathbf{a}}^s$ to get the final normalized centrality-based weights of the hybrid representations:
\begin{align}
    \mathbf{a} 
               &= [a^w_1, ..., a^w_m, a^s_1, ..., a^s_M], \\
    a^w_j &= \bar{a}^w_j / sum([\bar{\mathbf{a}}^w; \bar{\mathbf{a}}^s]), \\
    a^s_i &= \bar{a}^s_i / sum([\bar{\mathbf{a}}^w; \bar{\mathbf{a}}^s]),
\end{align}
where ``$[\cdot;\cdot]$'' represents concatenation.

Based on the hybrid representations (i.e., $\mathbf{X}$ and $\mathbf{R}_k$) and the centrality-based weights of the pseudo reference tokens and sentences (i.e., $\mathbf{a}$), we compute the relevance score between the summary and the pseudo reference by a weighted BERTScore~\cite{DBLP:conf/iclr/ZhangKWWA20_bertscore}. For brevity, we denote the $j$-th element of $\mathbf{X}$ as $\mathbf{x}_j$, the $i$-th element of $\mathbf{R}_k$ as $\mathbf{r}_i$, and the $i$-th element of $\mathbf{a}$ as $a_i$:
\begin{align}
    Recall &= \frac{\sum_i a_i \max_j \text{Sim}(\mathbf{r}_i, \mathbf{x}_j)}{\sum_i a_i}, \\
    Precision &= \frac{\sum_j \max_i \text{Sim}(\mathbf{r}_i, \mathbf{x}_j)}{|\mathbf{X}|}, \\
    F_1 &= \frac{2*Recall*Precision}{Recall + Precision}, 
\end{align}
where ``Sim'' denotes the cosine similarity and $|\mathbf{X}|$ equals to $n+N$. $Recall$, $Precision$, and $F_1$ are in the range of [-1, 1].

Besides the $F_1$ version, we also propose an adaptive $F_\beta$ version of relevance score as follows:
\begin{align}
    F_\beta &= \frac{(1 + \beta^2)*Recall*Precision}{Recall + \beta^2*Precision}, \\
    \beta^2 &= 
              \begin{cases}
              1, & \text{if}\ (\frac{|\mathbf{R}_k|}{|\mathbf{X}|})^{1/\gamma} \leq 1 \\
              2, & \text{if}\ (\frac{|\mathbf{R}_k|}{|\mathbf{X}|})^{1/\gamma} \geq 2 \\
              (\frac{|\mathbf{R}_k|}{|\mathbf{X}|})^{1/\gamma}, & \text{otherwise}
              \end{cases}, \label{equation: beta_square}
\end{align}
where $|\mathbf{R}_k|=m+M$, $|\mathbf{X}|=n+N$, and $\gamma$ is a positive integer hyper-parameter. In our experiments, $\gamma$ is set as 2 after fine-tuning on the validation dataset and is fixed for all the testing datasets. The physical meaning of $\beta$ is that the $Recall$ score is considered $\beta$ times as important as the $Precision$ score. In summarization evaluation, the coverage of the key information is always the most important quality indicator of the summary. Thus, we set the lower bound of $\beta$ as 1. On the other hand, the metric should not only evaluate the key information coverage, containing less unimportant content in the summary should also be considered. Therefore, we set the upper bound of $\beta$ as $\sqrt{2}$. As shown in Eq.\ref{equation: beta_square}, within the range of $[1, \sqrt{2}]$, $\beta$ adaptively changes according to the ratio between $|\mathbf{R}_k|$ and $|\mathbf{X}|$. 
The intuition comes from that a longer pseudo reference implies more key information needs to be covered by the summary. Besides, a shorter summary can easily get high precision but covers very limited important information in the pseudo reference. Thus, we give $Recall$ a higher weight to punish such short summaries when the pseudo reference is long. 

\smallskip
\noindent \textbf{Final Averaged Relevance Score.} After computing the centrality-weighted relevance score between the summary and the pseudo reference of each source document, we employ the average as the final relevance score of the summary:
\begin{align}
    score_{rel} = mean([F_*^1, ..., F_*^k, ..., F_*^K]),
\end{align}
where * is 1 for the $F_1$ variant and $\beta$ for the $F_\beta$ variant. The superscript $k$ indicates the $F_*$ score is computed with the $k$-th document. Note that $score_{rel} \in [-1, 1]$ and higher is better.

\subsection{Self-referenced Redundancy Score}
In this section, we introduce our self-referenced redundancy score. We engage the summary itself as the reference to evaluate the degree of the semantic similarity between each summary token or sentence with the other tokens or sentences. The averaged semantic similarity degree is used as the redundancy score. The computation is based on a self-masked BERTScore as follows:
\begin{align}
    score_{red} &= \frac{\sum_i \max_{j:i \neq j} \text{Sim}(\mathbf{x}_j, \mathbf{x}_i)}{|\mathbf{X}|}, \label{equation:redund}
\end{align}
where ``$j:i \neq j$'' means we do not consider the similarity between $\mathbf{x}_i$ and itself, i.e, self-masked. Because of the symmetric property, the $F_1$, precision, and recall scores are equal with each other. This is also the reason that we use precision in Eq.\ref{equation:redund} as the final redundancy score.  Note that $score_{red} \in [-1, 1]$ and lower is better.

\subsection{Final Evaluation Score}
After obtaining the relevance score and the redundancy score, we apply a linear combination to produce the final evaluation score of the summary based on the document set:
\begin{align}
    score = \frac{score_{rel} - \lambda * score_{red}}{1 + \lambda},
\end{align}
where $ 0 < \lambda \leq 1$ is a hyper-parameter to scale the redundancy score and $score \in [-1, 1]$. Higher $score$ means better summary quality. In our experiments, after fine-tuning on the validation set, $\lambda$ is set as 0.6 and is fixed for all the testing datasets. We denote the variants of our final method as \textbf{Ours($F_\beta$)-PacSumTopM} and \textbf{Ours($F_1$)-PacSumTopM} depending on whether the adaptive $F_\beta$ is employed.

\section{Experiment Setup}
\subsection{Datasets}
For comprehensively investigating our summarization evaluation methods, we test our methods on both multi-document and single-document summarization datasets. We leverage  TAC\footnote{https://tac.nist.gov/} datasets for multi-document summarization evaluation testing. We choose TAC-2010 as the validation dataset and TAC-2008/TAC-2009/TAC-2011 as the testing datasets. Following previous work, we only utilize the initial summaries in TAC datasets, i.e., the summaries for the document set A. For the single-document summarization evaluation, we employ CNNDM\footnote{https://bit.ly/price-of-debiasing}~\cite{chaganty-etal-2018-price-tmp} as the testing dataset.  The statistics of these datasets are shown in Table~\ref{table: dataset_statistics}. Note that the hyper-parameters of our methods are fine-tuned on TAC-2010 and then fixed for all the testing datasets.

For TAC datasets, we compute correlation coefficients between predicted scores of an evaluation method and the annotated Pyramid scores of summaries to measure the effectiveness of the method. Following \citet{DBLP:conf/acl/GaoZE20_supert}, a correlation is computed for each topic. Then, the averaged correlation from all the topics is engaged as the final correlation of the method with human ratings.

For CNNDM dataset, correlations are calculated with the human scores in three dimensions including \textit{Overall}, \textit{Grammar}, and \textit{Redundancy}. Following \citet{DBLP:journals/corr/abs-2010-01781_ls_score}, the correlation is computed between predicted scores of the $499 \times 4 = 1996$ (document, summary) pairs with corresponding human ratings.

\begin{table}[t]
    \centering
    \resizebox{\columnwidth}{!}{
    \begin{tabular}{|c| c|c| c c c| c c c|}
    \hline
    \multicolumn{2}{|c|}{\multirow{2}{*}{\textbf{Dataset}}} & \multirow{2}{*}{$|Topic|$} &\multicolumn{3}{c|}{\textbf{Document}} &\multicolumn{3}{c|}{\textbf{Summary}} \\ \cline{4-9}
     \multicolumn{2}{|c|}{} & & $|Set|$ & \textit{Ave.S} & \textit{Ave.T} &$|Systems|$ & \textit{Ave.S} & \textit{Ave.T}\\
    \hline
    \textit{Valid.} & \textbf{TAC-2010} & 46  & 10  & 23.2 & 651.8 & 43 & 4.3 & 118.9\\
    \hline
    \multirow{4}{*}{\textit{Test.}} 
    & \textbf{TAC-2011} & 44 & 10 & 20.1 & 560.5 & 50 & 4.3 & 120.9\\
    & \textbf{TAC-2009} & 44 & 10 & 24.9 & 705.8 & 55 & 4.1 & 117.6\\
    & \textbf{TAC-2008} & 48 & 10 & 23.3 & 660.0 & 58 & 4.2 & 119.6\\\cline{2-9}
    & \textbf{CNNDM} & 499 & 1 & 36.0 & 921.1 & 4 & 3.5 & 73.2\\
    \hline
    \end{tabular}
    }
    \caption{Statistics of datasets. ``\textit{Valid.}'' and ``\textit{Test.}'' indicate the dataset is used for validation and testing, respectively. ``$|Topic|$'' is the number of topics. Under each topic, a set of documents is given and summaries are from different systems associating with human-annotated quality scores. ``$|Set|$'' is the number of documents in the document set. ``\textit{Ave.S}'' and ``\textit{Ave.T}'' represent the averaged sentence number and token number per document or summary. Note that the token number is counted after the tokenization.
    ``$|Systems|$'' denotes the number of summarization systems in the dataset.
    }
    \label{table: dataset_statistics}
\end{table}

\begin{table*}[t]
\centering
\resizebox{0.72\textwidth}{!}{
\begin{tabular}{ | l | c c c | c c c | c c c |}
\hline
\multicolumn{1}{|c|}{\multirow{2}{*}{\textbf{Method}}} & \multicolumn{3}{c|}{\textbf{TAC-2011}} & \multicolumn{3}{c|}{\textbf{TAC-2009}} & \multicolumn{3}{c|}{\textbf{TAC-2008}} \\ \cline{2-10}
\multicolumn{1}{|c|}{}   & $r$    & $\rho$   & $\tau$ & $r$    & $\rho$   & $\tau$ & $r$    & $\rho$   & $\tau$ \\
\hline \hline

TF-IDF
& 0.313 & 0.294 & 0.209
& 0.372 & 0.382 & 0.279
& 0.375 & 0.341 & 0.243 \\

JS
& 0.377 & 0.333 & 0.240
& 0.376 & 0.381 & 0.279
& 0.385 & 0.338 & 0.242 \\

REAPER
& 0.377 & 0.334 & 0.237
& 0.358 & 0.357 & 0.256
& 0.287 & 0.261 & 0.187 \\



Ours($F_1$)-All
& 0.495	& 0.451	& 0.329
& 0.478	& 0.476	& 0.353
& 0.466	& 0.426	& 0.310 \\

Ours($F_\beta$)-All
& 0.498	& 0.455	& 0.332
& 0.480	& 0.471	& 0.348
& 0.462	& 0.423	& 0.307 \\

\hline


ROUGE-1-PacSumTopM
& 0.436	& 0.377	& 0.274
& 0.418	& 0.406	& 0.301
& 0.397	& 0.348	& 0.252 \\


ROUGE-2-PacSumTopM
& 0.429	& 0.388	& 0.287
& 0.380	& 0.419	& 0.314
& 0.410	& 0.355	& 0.259 \\


ROUGE-L-PacSumTopM
& 0.436	& 0.370	& 0.272
& 0.427	& 0.415	& 0.306
& 0.385	& 0.336	& 0.245 \\


MoverScore-PacSumTopM
& 0.521	& 0.475	& 0.351
& 0.483	& 0.485	& 0.362
& 0.479	& 0.440	& 0.323 \\


S+WMS-PacSumTopM
& 0.291	& 0.292	& 0.211
& 0.350	& 0.358	& 0.264
& 0.364	& 0.358	& 0.260 \\


C-ELMO-PacSumTopM
& 0.386	& 0.302	& 0.217
& 0.317	& 0.235	& 0.167
& 0.210	& 0.162	& 0.114 \\


C-SBERT-PacSumTopM
& 0.332	& 0.293	& 0.207
& 0.314	& 0.277	& 0.197
& 0.183	& 0.196	& 0.143 \\


SUPERT-PacSumTopM
& 0.511 & 0.481 & 0.357
& 0.486 & 0.494 & 0.368
& 0.493 & 0.457 & 0.334 \\

SUPERT-IDF-PacSumTopM
& 0.507	& 0.476	& 0.353
& 0.485	& 0.492	& 0.367
& 0.489	& 0.450	& 0.328 \\

\hline




Ours($F_1$)-PacSumTopM
& \underline{0.531}	& \underline{0.493}	& \underline{0.365}
& \underline{0.502}	& \underline{0.506}	& \textbf{0.381}		
& \underline{0.495}	& \underline{0.461}	& \underline{0.337} \\

Ours($F_\beta$)-PacSumTopM
& \textbf{0.541}	& \textbf{0.505}	& \textbf{0.374}
& \textbf{0.507}	& \textbf{0.508}	& \underline{0.380}
& \textbf{0.500}	& \textbf{0.465}	& \textbf{0.339} \\

\hline
\end{tabular}
}
\caption{Main results on multi-document summarization datasets. Pearson's $r$, Spearman's $\rho$, and Kendall's $\tau$ with human scores are reported. The best results are bold and the second-best results are underlined.}
\label{table: main-results-multi-doc}
\end{table*}

\begin{table*}[t]
\centering
\resizebox{0.72\textwidth}{!}{
\begin{tabular}{ | l | c c c | c c c | c c c |}
\hline
\multicolumn{1}{|c|}{\multirow{2}{*}{\textbf{Method}}} & \multicolumn{3}{c|}{\textbf{Overall}} & \multicolumn{3}{c|}{\textbf{Grammar}} & \multicolumn{3}{c|}{\textbf{Redundancy}} \\ \cline{2-10}
\multicolumn{1}{|c|}{}   & $r$    & $\rho$   & $\tau$ & $r$    & $\rho$   & $\tau$ & $r$    & $\rho$   & $\tau$ \\
\hline \hline

TF-IDF
& 0.264 & 0.249 & 0.187
& 0.186 & 0.170 & 0.127
& 0.281 & 0.253 & 0.187 \\

JS
& 0.265 & 0.232 & 0.174
& 0.210 & 0.180 & 0.136
& 0.317 & 0.278 & 0.208 \\

REAPER
& 0.036 & 0.032 & 0.024
& 0.004 & -0.006 & -0.005
& -0.020 & -0.031 & -0.024 \\

LS\_Score~\cite{DBLP:journals/corr/abs-2010-01781_ls_score}
& $-$	& 0.334 & $-$
& $-$	& 0.266 & $-$
& $-$	& 0.288 & $-$ \\



Ours($F_1$)-All
& 0.390	& 0.370	& 0.281
& 0.306	& 0.306	& 0.232
& 0.413	& 0.381	& 0.287 \\

Ours($F_\beta$)-All
& 0.361	& 0.337	& 0.255
& 0.273	& 0.270	& 0.204
& 0.395	& 0.356	& 0.268 \\

\hline


ROUGE-1-PacSumTopM
& 0.224	& 0.215	& 0.159
& 0.126	& 0.114	& 0.084
& 0.289	& 0.254	& 0.186 \\


ROUGE-2-PacSumTopM
& 0.347	& 0.335	& 0.253
& 0.254	& 0.240	& 0.181
& 0.398	& 0.369	& 0.274 \\


ROUGE-L-PacSumTopM
& 0.235	& 0.224	& 0.166
& 0.135	& 0.122	& 0.090
& 0.300	& 0.264	& 0.193 \\


MoverScore-PacSumTopM
& 0.373	& 0.341	& 0.259
& 0.264	& 0.240	& 0.181
& 0.411	& 0.359	& 0.267 \\


S+WMS-PacSumTopM
& 0.324	& 0.353	& 0.267
& 0.240	& 0.256	& 0.193
& 0.360	& 0.385	& 0.286 \\


C-ELMO-PacSumTopM
& 0.355	& 0.297	& 0.223
& 0.232	& 0.201	& 0.151
& 0.425	& 0.354	& 0.262 \\


C-SBERT-PacSumTopM
& \underline{0.405}	& 0.378	& 0.286
& 0.295	& 0.299	& 0.225
& 0.415	& 0.373	& 0.279 \\


SUPERT-PacSumTopM
& 0.384 & 0.374 & 0.284
& \underline{0.318} & \underline{0.317} & \underline{0.240}
& 0.381 & 0.369 & 0.277 \\

SUPERT-IDF-PacSumTopM
& 0.382	& 0.373	& 0.283
& 0.316	& 0.314	& 0.238
& 0.377	& 0.365	& 0.274 \\

\hline





Ours($F_1$)-PacSumTopM
& \textbf{0.416}	& \textbf{0.404}	& \textbf{0.308}
& \textbf{0.341}	& \textbf{0.341}	& \textbf{0.259}
& \textbf{0.428}	& \textbf{0.408}	& \textbf{0.308}\\

Ours($F_\beta$)-PacSumTopM
& 0.400	& \underline{0.381}	& \underline{0.290}
& 0.314	& 0.311	& 0.235
& \underline{0.427}	& \underline{0.395}	& \underline{0.298}\\

\hline
\end{tabular}
}
\caption{Main results on single-document summarization dataset (CNNDM).  Pearson's $r$, Spearman's $\rho$, and Kendall's $\tau$ with human scores are reported. The best results are bold and the second-best results are underlined.}
\label{table: main-results-single-doc}
\vspace{-0.15in}
\end{table*}

\subsection{Baselines}
In this section, we briefly introduce our baselines.

We choose \textbf{TF-IDF}, \textbf{JS}~\cite{DBLP:journals/coling/LouisN13_js}, and \textbf{REPEAR}~\cite{rioux-etal-2014-fear_reaper} as traditional reference-free baselines. 
All these traditional baselines do not build pseudo references and directly utilize the full content of the documents. For fairness, we also show the performance of our methods without building pseudo reference. We denote them as \textbf{Ours($F_1$)-All} and \textbf{Ours($F_\beta$)-All} since they use the whole document as a reference.

We also extend several popular reference-based methods as baselines. We adapt \textbf{ROUGE-1/2/L}~\cite{lin-2004-rouge-tmp}, \textbf{MoverScore}~\cite{zhao-etal-2019-moverscore-tmp}, and \textbf{S+WMS}~\cite{DBLP:conf/acl/ClarkCS19_s_wms} into the reference-free scenario via building the pseudo reference with the PacSumTopM method. We add the suffix ``\textbf{-PacSumTopM}'' to these baseline names to indicate the pseudo reference building process.

Besides, the SOTA reference-free summary evaluation metrics are also selected as our strong baselines, including \textbf{C-ELMO/C-SBERT}~\cite{sun-nenkova-2019-feasibility_c_elmo}, \textbf{SUPERT/SUPERT-IDF}~\cite{DBLP:conf/acl/GaoZE20_supert}, and 
\textbf{LS\_Score}~\cite{DBLP:journals/corr/abs-2010-01781_ls_score}. C-ELMO (C-SBERT) encodes the document and the summary using the pre-trained ELMO (SBERT) and then computes their cosine similarity. SUPERT-IDF is an extension of SUPERT, which utilizes the inverse document frequency (IDF) as the importance weight of each token. For fair comparisons, we also apply the same pseudo reference building process i.e., PacSumTopM, to  C-ELMO/C-SBERT/SUPERT/SUPERT-IDF and add the suffix ``\textbf{-PacSumTopM}'' to the their names.

\section{Results and Analysis}
\subsection{Main Results}
The main experimental results on multi-document summarization datasets are shown in Table~\ref{table: main-results-multi-doc}. We find that our $F_1$ version (i.e., Ours($F_1$)-PacSumTopM) already consistently outperforms all the baselines, which indicates the effectiveness of our centrality-weighted relevance score and our self-referenced redundancy score. The results also demonstrate that our $F_\beta$ version can further improve the performance of multi-document summarization evaluation. By comparing Ours($F_\beta$)-PacSumTopM and Ours($F_\beta$)-All, we see that the pseudo reference building process can significantly improve the performance. This is also the reason why we apply the same pseudo reference building process into SOTA baselines for fair comparisons. In the remaining part of this paper, we omit the suffix ``-PacSumTopM'' for simplicity when we mention a method.

\begin{figure}[t]
  \centering
  \includegraphics[width=\columnwidth]{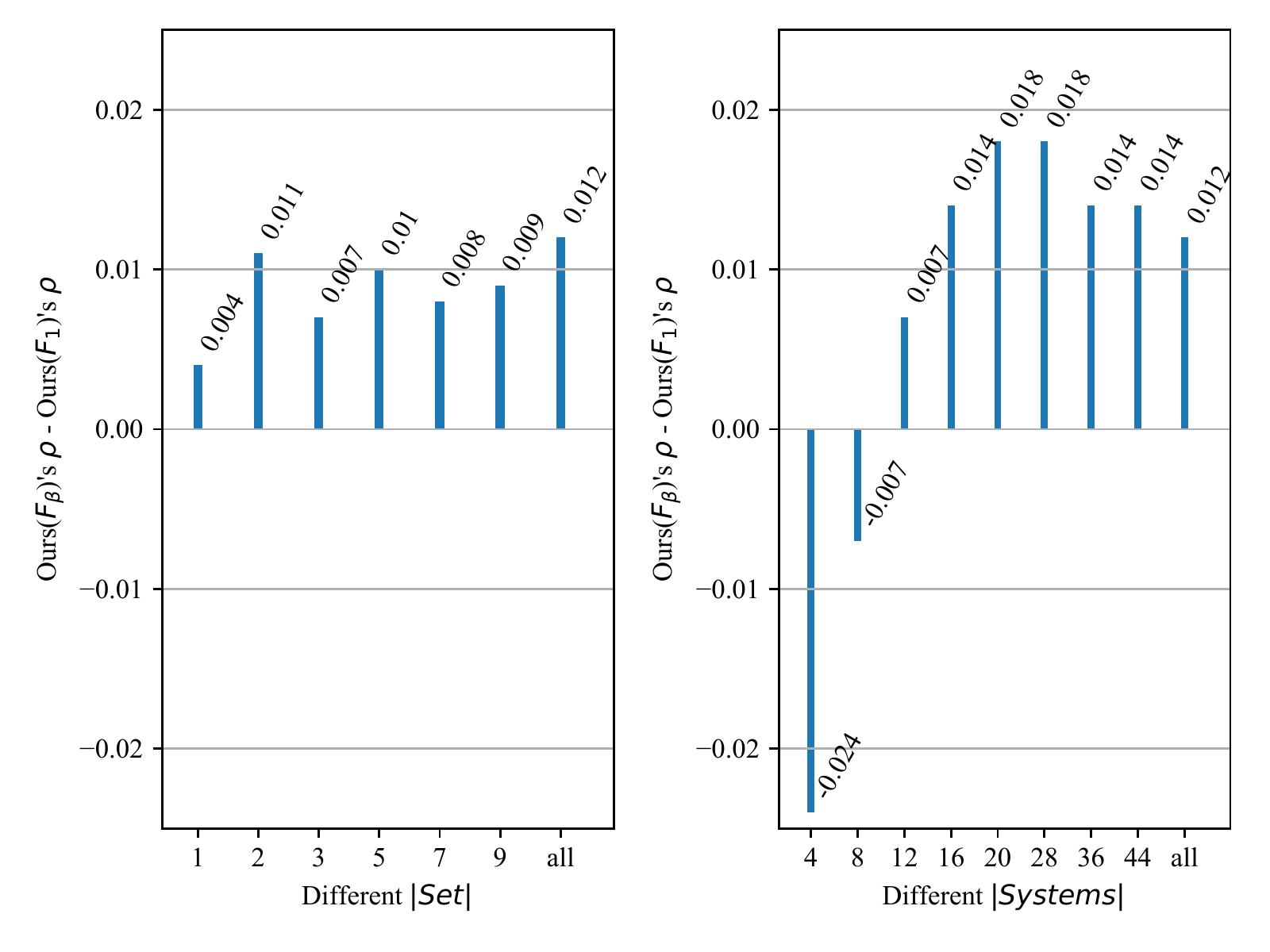}
  \vspace{-0.2in}
  \caption{The gap of Spearman's $\rho$ between Ours($F_\beta$) and Ours($F_1$) on TAC-2011 for different $|Set|$ and $|Systems|$. Positive gaps mean our $F_\beta$ can improve the performance while negative gaps indicate our $F_\beta$ degrades the performance. When changing one of them, the other is fixed. ``all'' means the full size is applied, i.e., 10 for $|Set|$ and 50 for $|Systems|$.}
  \label{figure: tac_mimic_cnndm}
  \vspace{-0.15in}
\end{figure}

We also test our methods on the single-document summarization dataset without further fine-tuning the hyper-parameters. The main results are displayed in Table~\ref{table: main-results-single-doc}. We note that our $F_1$ version still outperforms all the baselines, which manifests the high generalization ability of our $F_1$-based method. One interesting finding is that the performance significantly drops after incorporating the $F_\beta$ score.

To study the reason for the performance degradation on CNNDM after incorporating $F_\beta$, we compare CNNDM and TAC datasets first. From Table~\ref{table: dataset_statistics}, we note the main differences between them are the size of the document set for each topic (i.e., $|Set|$) and the number of the summarization systems (i.e., $|Systems|$). CNNDM has much smaller $|Set|$ and $|Systems|$. We use the TAC-2011 dataset as an example to investigate whether our $F_\beta$ is unsuitable for smaller $|Set|$ and $|Systems|$. We change $|Set|$ and $|Systems|$ respectively and report the gap of Spearman's $\rho$ between Ours($F_\beta$) and Ours($F_1$) in Figure~\ref{figure: tac_mimic_cnndm}. From the results, we observe that our $F_\beta$ can consistently improve the performance for different $|Set|$. For the single-document summarization setting, i.e., $|Set|$=1, it still obtains a positive gap. Nevertheless, when the  $|Systems|$ is small such as 4, applying our $F_\beta$ leads to a dramatic performance dropping. From Table~\ref{table: dataset_statistics}, we also see that CNNDM and TAC-2011 have different summary lengths (73.2 for CNNDM and 120.9 for TAC-2011). However, when we limit the $|Systems|$ of TAC-2011 to smaller numbers, the average length of generated summaries is still around 120, which indicates the performance degeneration is indeed from the change of system numbers. Therefore, we suggest using Ours($F_\beta$) when $|Systems|$ is large like 12 and employing Ours($F_1$) when $|Systems|$ is small like 4.

\subsection{Ablation Study}
\begin{figure}[t]
  \centering
  \includegraphics[width=\columnwidth]{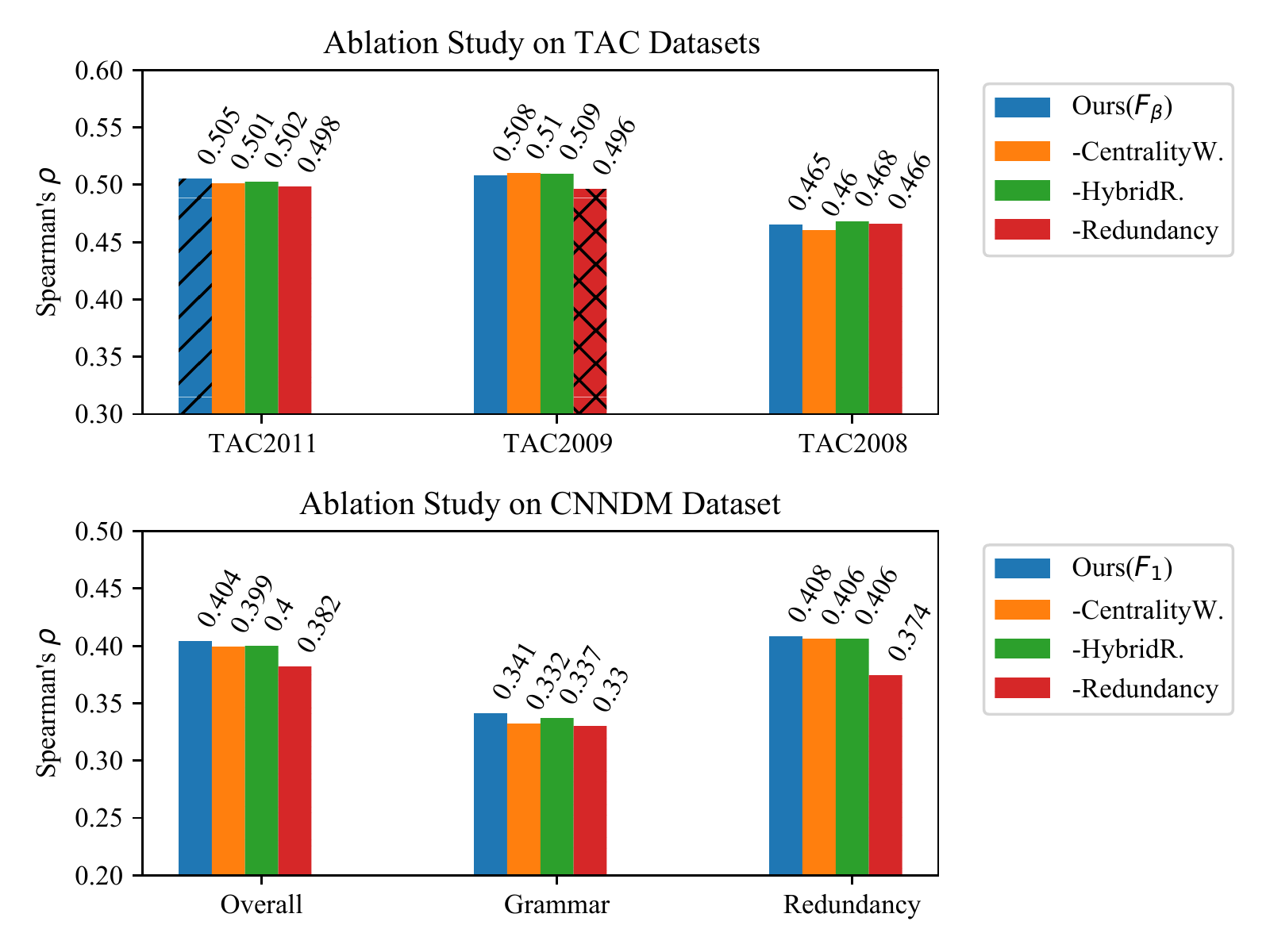}
  \caption{Ablation studies for Ours($F_\beta$) on TAC datasets and Ours($F_1$) on CNNDM. ``-CentralityW.'' means that we remove the centrality weighting when computing relevance scores. ``-HybridR.'' represents we only utilize the token-level representations when calculating relevance and redundancy scores. ``-Redundancy'' indicates we omit the redundancy score.} 
  \label{figure: ablation_study}
\end{figure}

For better understanding the contributions of our proposed components, we conduct ablation studies on the best-performed method on each dataset, i.e., Ours($F_\beta$) for the multi-document summarization datasets and Ours($F_1$) for the single-document summarization dataset. We display results of the rank-based Spearman's $\rho$ in Figure~\ref{figure: ablation_study}.

As shown in the figure, after removing one of the three components (i.e., the centrality weighting, the hybrid representation, and the redundancy score), the performance of our methods become worse in most cases. This finding demonstrates the effectiveness of our proposed components. Besides, we also note that removing the redundancy score significantly degrades the performance on the redundancy evaluation on CNNDM, which indicates our redundancy score effectively captures the redundancy degree of the summaries.

\begin{table}[t]
    \centering
    \resizebox{\columnwidth}{!}{
    \begin{tabular}{|l| c c c| c c c|}
    \hline
    \multirow{2}{*}{\textbf{Method}} &\multicolumn{3}{c|}{\textbf{TAC}} &\multicolumn{3}{c|}{\textbf{CNNDM}} \\ \cline{2-7}
     
    & \textbf{2011} & \textbf{2009} & \textbf{2008} 
    & \textbf{Overall} & \textbf{Grammar} & \textbf{Redundancy} \\
    \hline
    Ours($F_1$) & 0.493	& 0.506	& 0.461	& 0.404	& 0.341	& 0.408\\
    Ours($F_\beta$) & 0.505	& 0.508	& 0.465	& 0.381	& 0.311	& 0.395\\
    \hline
    MoverScore & 0.475	& 0.485	& 0.440	& 0.341	& 0.240	& 0.359 \\
    +CentralityW. & 0.472	& 0.467	& 0.431	& 0.350	& 0.257	& 0.364 \\
    +Redundancy & 0.237	& 0.202	& 0.221	& 0.448	& 0.326	& 0.546 \\
    +Both & 0.261	& 0.220	& 0.241	& 0.455	& 0.341	& 0.545 \\
    \hline
    \end{tabular}
    }
    \caption{Spearman's $\rho$ of incorporating the centrality weighting and redundancy score into MoverScore based framework. ``+Both'' means these two features are simultaneously applied.}
    \label{table: apply_to_moverscore}
    \vspace{-0.15in}
\end{table}

\subsection{Apply Centrality Weighting and Redundancy Score into MoverScore}
Besides basing on BERTScore, we also study whether our key features i.e., the centrality weighting and redundancy score, can work well in a MoverScore based framework (i.e., the relevance and redundancy scores are computed using MoverScore). Note that our $F_\beta$ is not applicable to MoverScore since it is not an $F$-measure. The results are listed in Table~\ref{table: apply_to_moverscore}. We find that these two features significantly improve the performance of the original MoverScore on single-document summarization evaluation while degrading the performance dramatically on multi-document summarization evaluation. On CNNDM, the enhanced MoverScore even outperforms Ours($F_1$) on the ``Overall'' and ``Redundancy'' aspects, which indicates MoverScore is a promising basis for our proposed new features. We leave solving the performance dropping of the enhanced MoverScore on multi-document setting as future work.

\subsection{Robustness Analysis}
\begin{figure}[t]
  \centering
  \includegraphics[width=\columnwidth]{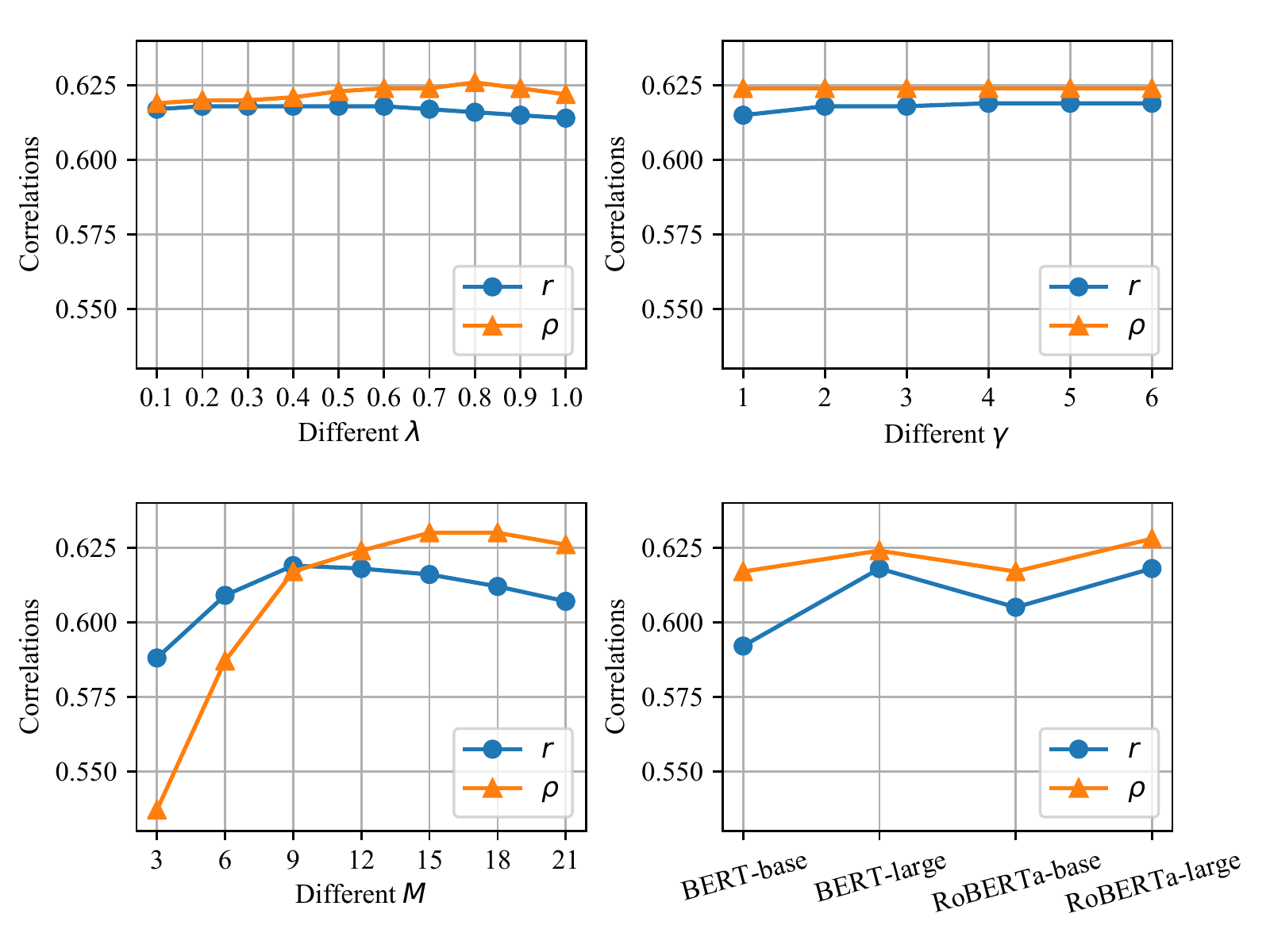}
  \caption{The performance of Ours($F_\beta$) on TAC-2010 under different $\lambda$, $\gamma$, $M$, and encoding models. When we change one of them, the others are fixed. The Pearson's $r$ and Spearman's $\rho$ are reported.}
  \label{figure: robustness_analysis}
\end{figure}

We investigate the robustness of our method on the following factors and report the experimental results on the validation dataset (i.e., TAC-2010)  in Figure~\ref{figure: robustness_analysis}:
(1) the hyper-parameter $\lambda$  for scaling the redundancy score; (2) the hyper-parameter $\gamma$  in $F_\beta$;  (3) the number of selected sentences for pseudo reference i.e., $M$; (4) different pre-trained contextual encoding models including BERT-base\footnote{bert-base-nli-stsb-mean-tokens}, BERT-large\footnote{bert-large-nli-stsb-mean-tokens}, RoBERTa-base\footnote{roberta-base-nli-stsb-mean-tokens}, and RoBERTa-large\footnote{roberta-large-nli-stsb-mean-tokens}.

Since both Spearman's $\rho$ and Kendall's $\tau$ are rank-based correlation coefficients, we omit Kendall's $\tau$ for simplicity. From this figure, we observe that the performance of our method is relatively stable for different $\lambda$ and $\gamma$. We also find that a small $M$ leads to lower correlations because much important information may be abandoned when building the pseudo references. But a large $M$ will also degenerate the correlations since more noises are introduced. Thus, a moderate $M$ is better. As for encoding models, we note that large encoding models obtain better performance than base encoding models. However, large models need more computation resources and time to encode the input text. Note that for our final method, we only fine-tune $\lambda$ and $\gamma$ on the TAC-2010 and set them as 0.6 and 2. As for $M$ and encoding models, following the configuration of  SUPERT~\cite{DBLP:conf/acl/GaoZE20_supert}, we directly set $M$ as 12 and employ the BERT-large as the encoding model. All these factors are fixed for all testing datasets.

\subsection{Performance on Bad/Good Summaries}
\begin{figure}[t]
  \centering
  \includegraphics[width=\columnwidth]{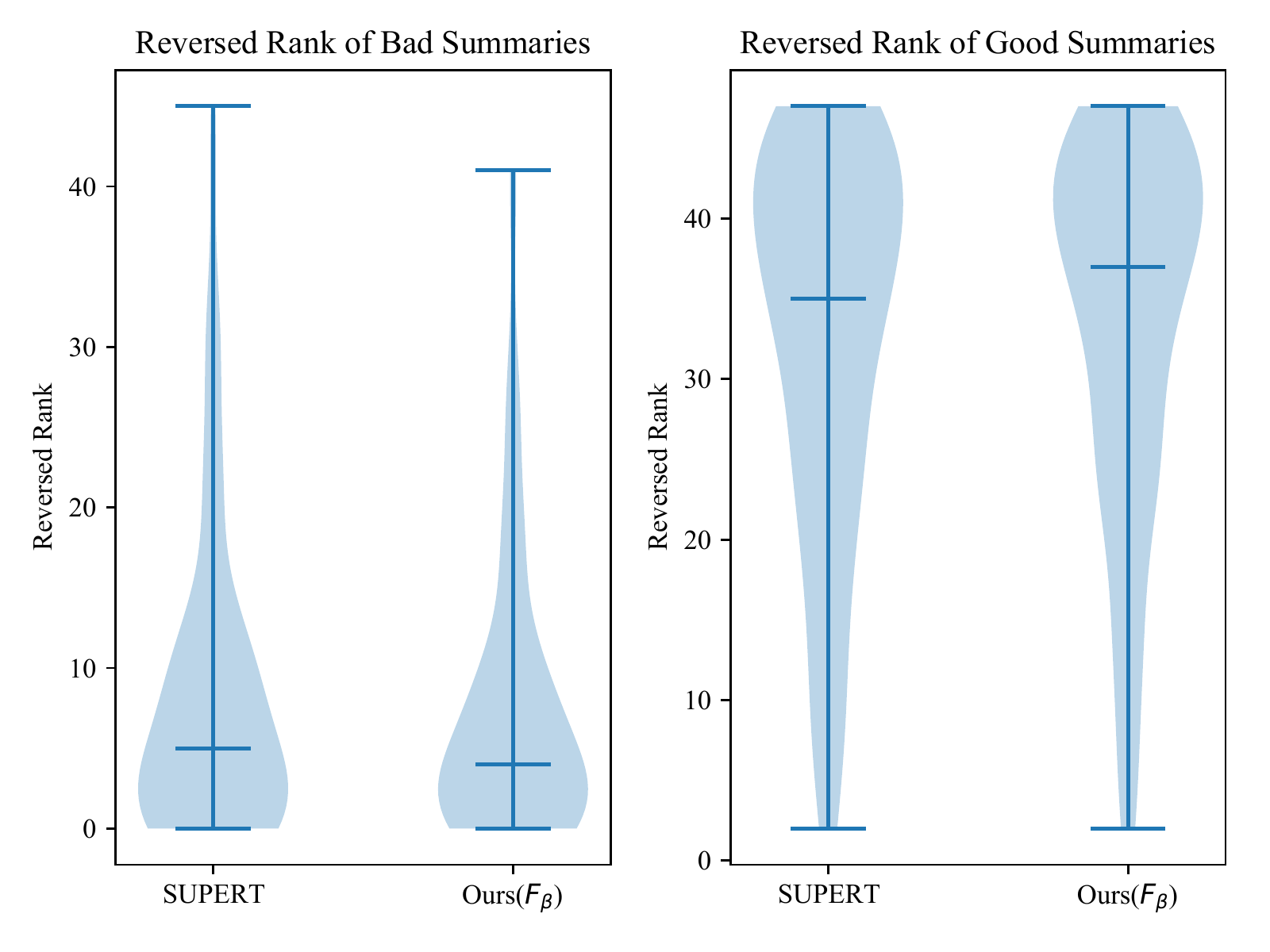}
  \caption{Distributions of the reversed rank from SUPERT and Ours($F_\beta$) for bad and good summaries on TAC-2011. The bar in the middle indicates the median.}
  \label{figure: bad_good_reverse_rank}
\end{figure}

In this section, we evaluate the ability of our method to distinguish bad and good summaries. The bad and good summaries are selected by human ratings. We use TAC-2011 as an example and choose SUPERT as a strong baseline. The corresponding distributions of the reversed rank for bad and good summaries are illustrated in Figure~\ref{figure: bad_good_reverse_rank}. 
A smaller (larger) reversed rank represents the summary is assigned with a lower (higher) score.
From the figure, we find that compared with SUPERT, Our($F_\beta$) has a better ability to assign bad summaries lower scores and good summaries higher scores, which demonstrates the effectiveness of our method again. Moreover, we also note that both SUPERT and Ours($F_\beta$) are good at giving bad summaries lower scores while having difficulty in assigning good summaries higher scores. We leave solving this problem as another future work under the reference-free setting.

\section{Related Work}
\smallskip
\noindent \textbf{Reference-based Evaluation Metrics} mainly measure the relevance between the human-annotated references and the system-generated text, which are widely adopted in text summarization~\cite{lin-2004-rouge-tmp, zhao-etal-2019-moverscore-tmp}, machine translation~\cite{papineni-etal-2002-bleu, DBLP:conf/iclr/ZhangKWWA20_bertscore}, and dialogue systems~\cite{papineni-etal-2002-bleu, gao2021ream, xiang2021assessing}. For example, ROUGE~\cite{lin-2004-rouge-tmp} evaluates the token sequence overlapping. BERTScore~\cite{DBLP:conf/iclr/ZhangKWWA20_bertscore}, S+WMS~\cite{DBLP:conf/acl/ClarkCS19_s_wms}, and MoverScore~\cite{zhao-etal-2019-moverscore-tmp} measure the semantic similarity between the references and the summary via a greedy or optimized minimum Earth Mover's Distance.

\smallskip
\noindent \textbf{Reference-free Evaluation Metrics} have been developed to avoid the dependency on human-annotated references, which obtain more and more attention in recent years~\cite{bohm-etal-2019-better, DBLP:conf/acl/GaoZE20_supert, DBLP:journals/corr/abs-2010-01781_ls_score,  chan2021enhancing}. Some of them need to train a scorer~\cite{DBLP:conf/naacl/PeyrardG18_statistic_training,xenouleas-etal-2019-sum_summQE,scialom-etal-2019-answers_QA,bohm-etal-2019-better}. For example, LS\_Score~\cite{DBLP:journals/corr/abs-2010-01781_ls_score} designs a metric which combines a linguistic quality scorer trained from the built positive and negative summaries, and a relevance scorer based on cosine similarity. The others do not require training~\cite{DBLP:journals/coling/LouisN13_js,rioux-etal-2014-fear_reaper,peyrard-2019-simple_theoretical,sun-nenkova-2019-feasibility_c_elmo}. For instance, SUPERT~\cite{DBLP:conf/acl/GaoZE20_supert} builds the pseudo references from the source document first and then engages BERTScore to compute the relevance score between the pseudo reference and the summary.

\section{Conclusion}
In this paper, we propose a novel training-free and reference-free summarization evaluation metric consisting of a relevance score and a redundancy score. Experiments on multi-document and single-document summarization settings show the effectiveness of our methods. One promising future direction is to solve the performance dropping issue after applying our key features into MoverScore and the other is to tackle the problem that current metrics struggle to assign higher scores for good summaries.

\section*{Acknowledgements}
The work described in this paper was partially supported by the Research Grants Council of the Hong Kong Special Administrative Region, China (CUHK 2410021, Research Impact Fund (RIF), R5034-18).

\bibliographystyle{acl_natbib}
\bibliography{acl2021}

\end{document}